# Machine learning applied to emerald gemstone grading: framework proposal and creation of a public dataset

F. B. Pena[1] · D. Crabi[1] · S. C. Izidoro[1] · É. O. Rodrigues[2] · G. Bernardes[1]

**Abstract**
The grading of gemstones is currently a manual procedure performed by gemologists. A popular approach uses reference stones, where those are visually inspected by specialists that decide which one of the available reference stone is the most similar to the inspected stone. This procedure is very subjective as different specialists may end up with different grading choices. This work proposes a complete framework that entails the image acquisition and goes up to the final stone categorization. The proposal is able to automate the entire process apart from including the stone in the created chamber for the image acquisition. It discards the subjective decisions made by specialists. This is the first work to propose a machine learning approach coupled with image processing techniques for emerald grading. The proposed framework achieves 98% of accuracy (correctly categorized stones), outperforming a deep learning approach. Furthermore, we also create and publish the used dataset that contains 192 images of emerald stones along with their extracted and pre-processed features.

## 1 Introduction

Gemstones worth nearly 22 billion dollars were sold in 2018 according to the Future Market Insights [9]. This market is also expected to grow by approximately 5% each year. Among the colored stones, emeralds are the most lucrative in the USA [11].

✉ G. Bernardes
giovanibernardes@unifei.edu.br

F. B. Pena
filipe.b.pena@gmail.com

D. Crabi
danilocrabi@gmail.com

S. C. Izidoro
sandroizidoro@unifei.edu.br

É. O. Rodrigues
erickrodrigues@utfpr.edu.br

1 Laboratory of Robotics, Intelligent and Complex Systems (RobSIC) - Institute of Science and Technology, Universidade Federal de Itajubá (UNIFEI), Itabira, State of Minas Gerais, Brazil

2 Department of Academic Informatics (DAINF), Universidade Tecnológica Federal do Paraná (UTFPR), Pato Branco, State of Parana, Brazil

Emerald, sapphire and ruby are the most desired colored gemstones. This “big three” is associated with more economic activity when compared to all the rest of the colored gemstones combined. The value of emeralds alone imported into the USA in 2015 surpassed the value of colored stones out of the big three altogether [11].

Value factors hinge largely on color, with nuances of saturation and hue affecting price to a significant degree. The most desirable color is a slight bluish green in a medium dark tone with strong to vivid saturation [5]. These properties are manually measured by a specialist, leaving a lot of room for automation. Furthermore, multiple factors such as ambient light, tools used for measuring, fatigue, age, overwork and intersubject perception can affect the grading and subsequent market price estimation [33].

This work proposes a framework based on an image acquisition protocol, feature extraction and machine learning to provide a categorical grading estimation for emeralds. This framework is fully automated. We also provide the first public emerald dataset in the literature (emerald stone images and extracted features) for further reproduction and comparison.

## 2 Literature review

In the gemstone industry, stone esthetics or beauty is inferred according to color, clarity, cut, and carat (weight). These four properties are commonly called the four Cs. Clarity is usually associated with the deposit of other materials within the gemstone such as other minerals and air bubbles. Color is associated with and can be decomposed into hue, saturation and brightness [5]. Weight and cut are straightforward measurements.

Emerald is the green variety of the beryl mineral. It is one of the most valuable precious gems after ruby and diamond. When it comes to the market, its color is more important than clarity and brilliance. In 2000, a 10.11 ct Colombian cut was sold for US$1,149,850 [13].

Emerald is considerably rare. However, it can be found on all five continents. Brazil, Colombia, Russia, Zambia, Zimbabwe, Madagascar, Afghanistan and Pakistan are the largest producers of emeralds [13]. Deposits of emeralds occur mainly in eastern Brazil, eastern Africa, south Africa, Madagascar, India, Australia, as well as in young volcano-sedimentary series in Canada, Pakistan, Bulgaria, Russia, Canada and Spain [13].

Pieces that vary from a pure green to a bluish green while also being bright and highly saturated are associated with a higher pricing. Sizes usually vary from a few millimeters to 1.5 mm. Tools such as pachymeters are used to measure the stones through a manual procedure, which requires attention and precision from the specialist. The specialist has to coordinate the pachymeter with tweezers [16].

Some works in the literature propose methods for measuring the quality of garnets in gemstones [18, 29]. However, these works are gemology-oriented and provide manual guidelines for a broad spectrum of gemstones. Similarly, the work [19] provides general guidelines for a manual analysis of emerald stones, specifically.

In terms of computer science and automation, the works of [31] and [28] use image processing to determine the grade of gemstones. Both works provide a complete hardware and software solution and separate their methodology in four steps: (1) image acquisition, (2) background extraction, (3) feature extraction and (4) categorization. Although effective, the work of [31] targets opala stones while [28] targets amber stones.

It is clear that an automated solution for emerald stones is nonexistent in the literature. This work contributes to the literature by proposing a novel solution based on image processing and machine learning while also providing a public dataset containing emerald stones. In what follows, we shift the attention toward the two main different types of learning.

Supervised learning consists of analyzing labeled instances in order to generate a classification model capable of producing labels for unlabeled instances [22, 24]. We are able to predict the category of new emerald stones after training with previously labeled emerald stones. Supervised learning is well established and vastly employed in fields such as computer science and engineering, and in a great number of sub-areas, such as robotics [7], computer vision [20, 25], data mining and knowledge discovery [4], health care [21, 27] and many others.

In terms of supervised learning, we use the multilayer perceptron algorithm [30], which is a neural network, the random forests [2] classifier, which is a powerful ensemble decision tree algorithm and the extremely randomized tree [12], which is also an ensemble approach based on decision trees.

Unsupervised learning, on the other hand, usually leans toward a grouping or a clustering approach [15], although other unsupervised approaches do exist in the literature. Clustering is not preceded by training phases and does not have access to the label or class information. Clustering algorithms are capable of grouping the stone images according to their features, which are extracted from statistical values usually computed from edge detection and texture data. The main difference lies on the fact that clustering methods are amenable to the inclusion of new stone classes in the dataset. However, they are slower to compute as no previously trained model is created.

In terms clustering, we use the k-Means algorithm [1], which is the most popular and trivial clustering algorithm. k-Means starts off with randomized cluster centers, associates all classes to clusters and recalculates the cluster centers. This process is repeated several times until convergence. We also use the affinity propagation algorithm [32] that unlike k-Means does not require the paramater $k$, which determines the number of clusters.

In what follows, we describe the proposed methodology, which uses a pre-processing step to extract the emerald stone from the background. Later, we apply clustering (unsupervised) and classification (supervised) learning. We compare the performance of both learning approaches in our dataset.

## 3 Materials and methods

In the context of machine learning, results rely heavily on the quality of the acquired images. Three aspects can lead to improvements in terms of machine learning recognition: (1) camera distance, (2) quality of the camera sensor, (3) constant and neutral ambient light. Cameras that are closer to the gemstone can capture smaller and finer details, which can potentially improve the precision of the extracted information and subsequently the results.

Similarly, cameras with better sensors also improve the acquisition of finer details. Constant light is preferred in order to maintain the same pattern over the entire image acquisition process. This reduces the introduced variation that the classifier or clusterization algorithm has to generalize with. The white light is important as colored lights would modify the perception of color registered by the camera.

Respecting the topics raised above, a small chamber was created to capture the images, standardizing the process while also avoiding external reflections. The chamber is a closed small crate that isolates the emerald stones from the external ambient light. The interior base of the chamber was painted white. Table 1 highlights the specifications of the camera used in this work. Figure 1 shows the created chamber.

The emerald stones were provided by a mining company based in Brazil. The usual grading process is based on reference stones. First, gemologists chose a few stones as reference. Each reference stone is placed at one of the narrow channels shown in Fig. 2. Later, emerald stones are analyzed and grouped with the most similar reference stone. This manual process using reference stones entails a brightness and color analysis.

At total, gemologists work with three possible brightness levels and five potential color levels. This sums to a total of 15 possibilities. The stones we were able to capture belong to a total of 8 out of the 15 possible categories. In other words, this dataset does not cover the entire spectrum of gradings or categories for emerald gemstones. Some of these categories are very rare and valuable and were absent from the dataset due to unavailability. We captured a total of 192 emerald stones, 24 images belonging to each of the 8 categories, well-balanced and distributed dataset. Figure 3 shows stones from each one of the 8 available categories (ranging from 0 to 7), and Table 2 shows all the categories available in the created dataset.

### 3.1 Proposed method

The method includes three overall steps: (1) preprocessing, which consists of removing the background from the emerald stone, (2) feature extraction, where texture and edge information are computed and extracted from the preprocessed image and (3) machine learning, where supervised and unsupervised learning algorithms are used to group and to classify new emerald stones based on the extracted features.

The Region of Interest (ROI) is obtained after a threshold operation followed by a morphological closing (morphological dilations followed by morphological erosions to remove the holes) [24]. These two operations produce a mask that is multiplied over the original image containing the emerald stones. This multiplication separates the stone from the background. Figure 4 highlights this process in order.

**Table 1** Camera specification

| Manufactuter | FLIR |
|---|---|
| Model | Blackfly S USB3 - BFS-U3-200S6C-C |
| Megapixels | 20 |
| Resolution | 5472 × 3648 |
| Sensor | Sony IMX183 |

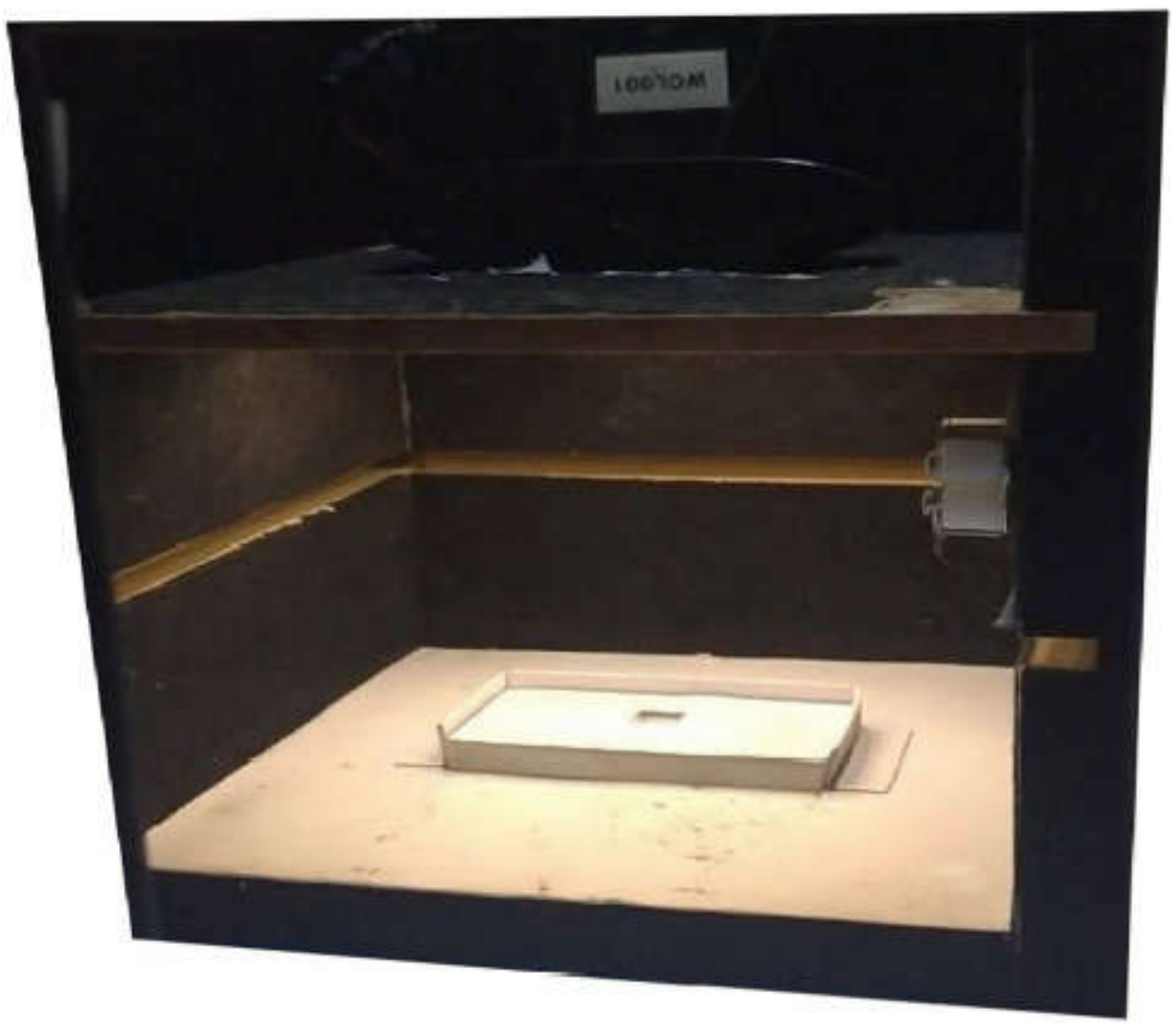

**Fig. 1** Chamber used for the image acquisition

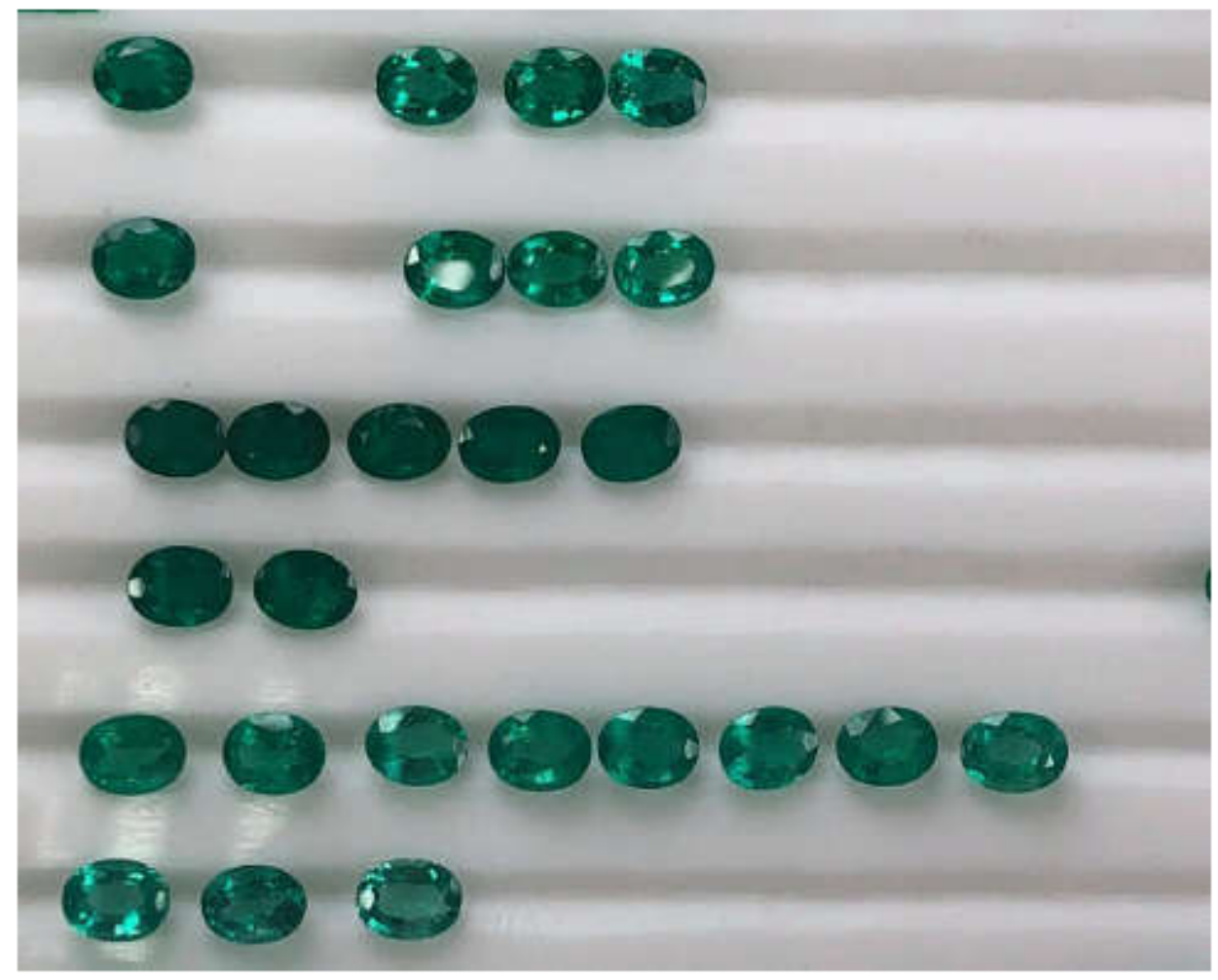

**Fig. 2** Manual grouping (association by reference) of emerald stones

**Table 2** Categories of the dataset

| | Color 1 | Color 2 | Color 3 | Color 4 | Color 5 |
|---|---|---|---|---|---|
| Brightness 1 | | Category 0 | Category 7 | | |
| Brightness 2 | Category 6 | Category 3 | Category 1 | Category 4 | Category 2 |
| Brightness 3 | | | | | Category 5 |

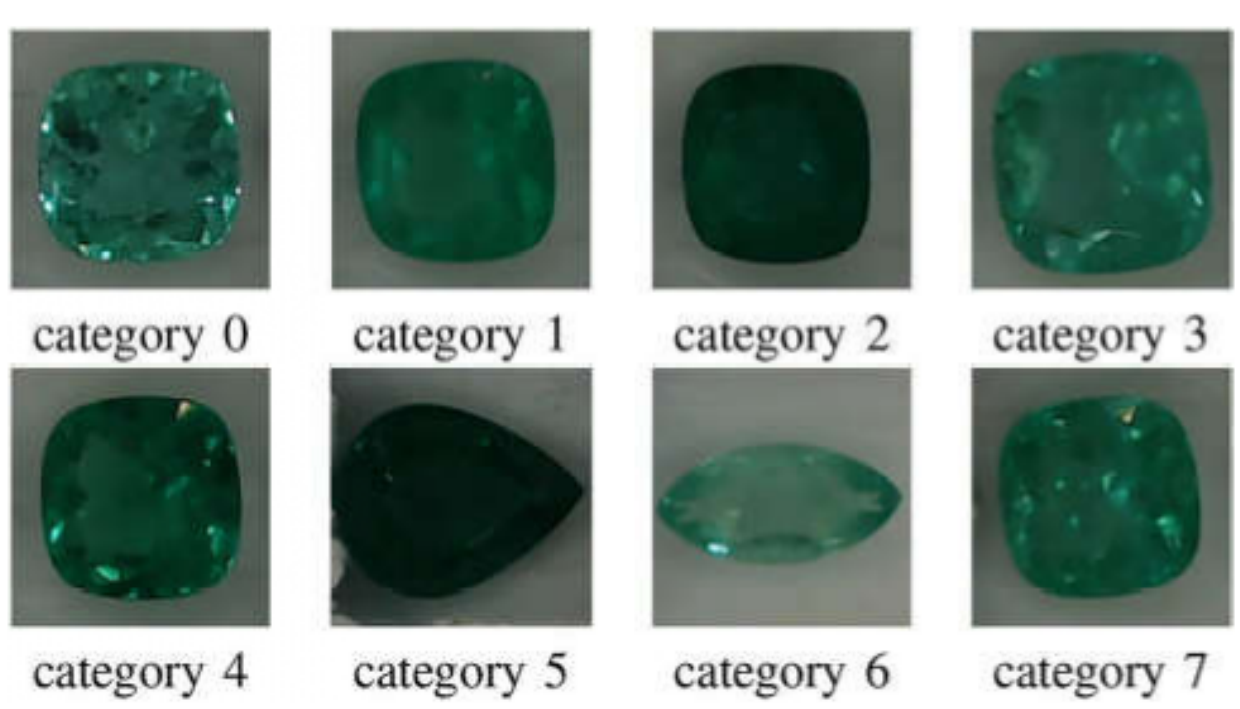

**Fig. 3** Emerald instances belonging to different categories found in the dataset

### 3.2 Feature extraction

A set of features is extracted from the images containing emerald stones, which is further used by the machine learning algorithms to determine or predict the category of the stones. We extracted a total of 24 statistical features. Among them, we compute some features from the gray-level co-occurrence matrix (GLCM or COM) [23, 25] such as homogeneity and entropy.

GLCMs contain the co-occurrences of pixel gray values at a given distance. Figure 5 shows an example of a GLCM. In this case, the chosen distance is (1, 0), which is analogous to the pixel at the right. The position (1, 2) of the matrix has the value 4 due to four co-occurrences respecting the distance (1, 0) of the values 1, 2 in the image (highlighted in red).

The co-occurrence matrix $C$ associates the number of co-occurrences of a gray level of a pixel $a$ to a gray level $b$ in an image $P$ at a given distance parameter $(\Delta x, \Delta y)$ for all pixels in $P$, as shown in Eq. 1:

$$C_{\Delta x,\Delta y}(a,b) = \sum_i \sum_j \begin{cases} 1, & if \;\; p_{i,j} = a \;\; and \;\; P_{i+\Delta y, j+\Delta x} = b \\ 0, & otherwise \end{cases} \tag{1}$$

where $i$ and $j$ represent positions in the input image $P$.

The homogeneity is computed based on the normalized GLCM as follows:

$$Homogeneity = \sum_{h,k} \frac{C_{\Delta x,\Delta y}(h,k)}{1+(h-k)^2} \tag{2}$$

where $h$ and $k$ are values in the GLCM.

The entropy is computed as follows:

$$Entropy = \sum_{h,k} ln(C_{\Delta x,\Delta y}(h,k))\, C_{\Delta x,\Delta y}(h,k) \tag{3}$$

**Fig. 4** ROI extraction process

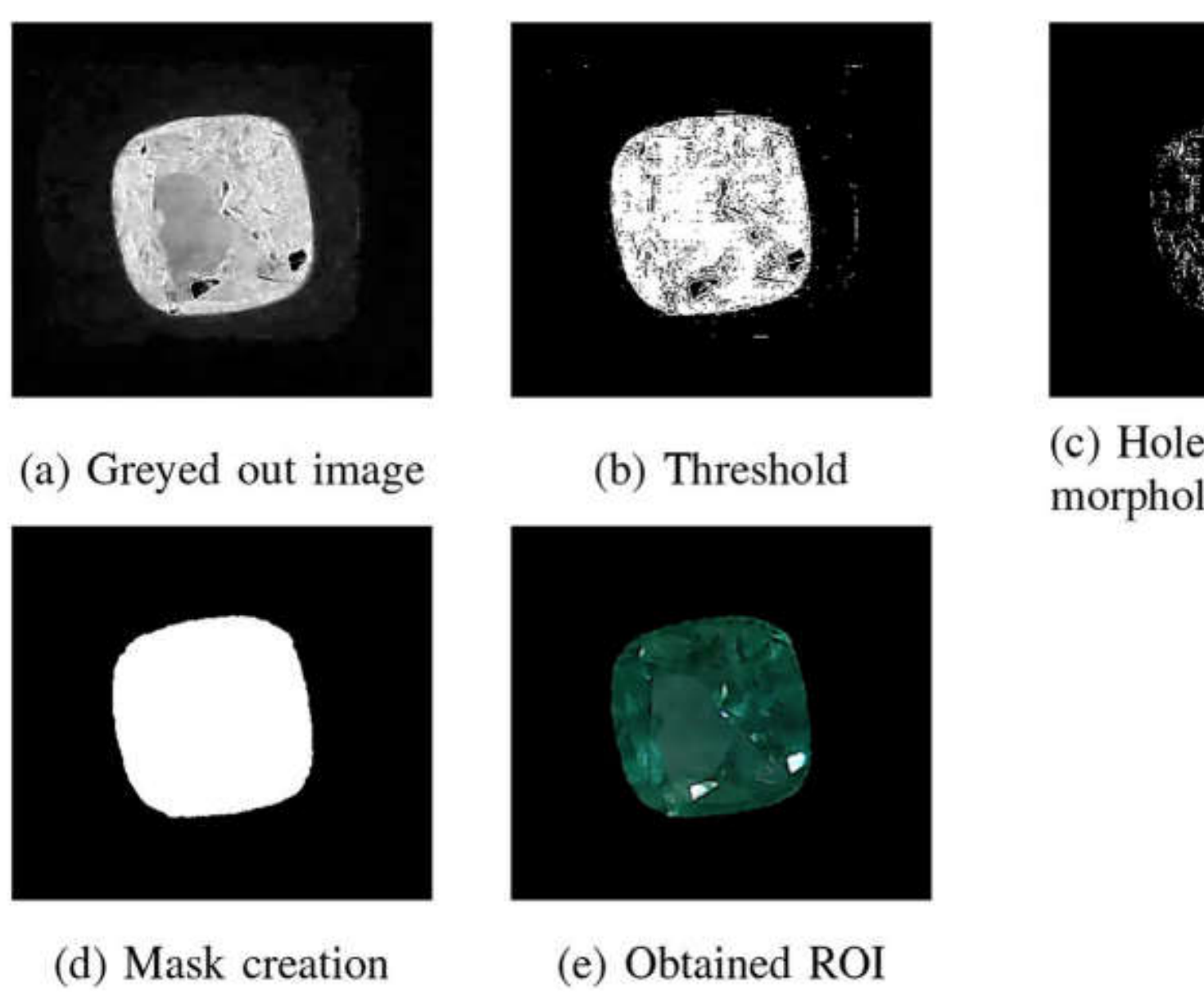

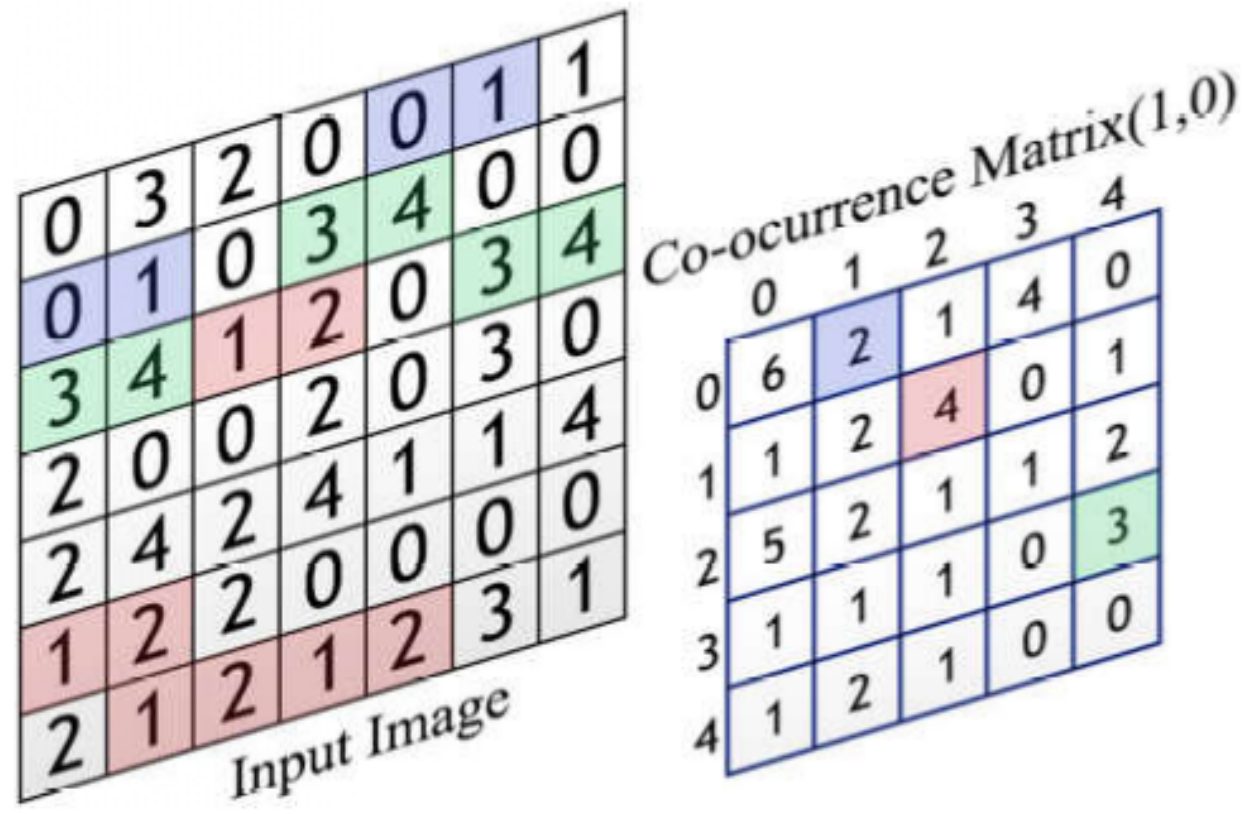

**Fig. 5** Example of a GLCM. The matrix is created from the data extracted from the input image at the left

We use the Hue-Saturation-Value (HSV) color space. Images are converted to this color space as it is easier to quantify brightness and color, in contrast to the usual Red-Green-Blue (RGB) color space.

The feature vector then contains a total of 24 features. The first four are very intuitive and were extracted directly from these HSV channels: (f1) arithmetic mean of the S channel, (f2) standard deviation of the S channel, (f3) arithmetic mean of the V channel, (f4) standard deviation of the V channel. From feature number 5 to feature number 20, we incorporate the idea of the reference stones that are used by specialists, as follows:

- f5: Histogram comparison using the Bhattacharrya distance (current image and reference image from category 0), this is computed for the S or saturation channel.
- f6: Histogram comparison using the Bhattacharrya distance (current image and reference image from category 1), this is computed for the S or saturation channel.
- f7: Histogram comparison using the Bhattacharrya distance (current image and reference image from category 2), this is computed for the S or saturation channel.
- f8: Histogram comparison using the Bhattacharrya distance (current image and reference image from category 3), this is computed for the S or saturation channel.
- f9: Histogram comparison using the Bhattacharrya distance (current image and reference image from category 4), this is computed for the S or saturation channel.
- f10: Histogram comparison using the Bhattacharrya distance (current image and reference image from category 5), this is computed for the S or saturation channel.
- f11: Histogram comparison using the Bhattacharrya distance (current image and reference image from category 6), this is computed for the S or saturation channel.
- f12: Histogram comparison using the Bhattacharrya distance (current image and reference image from category 7), this is computed for the S or saturation channel.
- f13: Histogram comparison using the Bhattacharrya distance (current image and reference image from category 0), this is computed for the V or value channel.
- f14: Histogram comparison using the Bhattacharrya distance (current image and reference image from category 1), this is computed for the V or value channel.
- f15: Histogram comparison using the Bhattacharrya distance (current image and reference image from category 2), this is computed for the V or value channel.
- f16: Histogram comparison using the Bhattacharrya distance (current image and reference image from category 3), this is computed for the V or value channel.
- f17: Histogram comparison using the Bhattacharrya distance (current image and reference image from category 4), this is computed for the V or value channel.
- f18: Histogram comparison using the Bhattacharrya distance (current image and reference image from category 5), this is computed for the V or value channel.
- f19: Histogram comparison using the Bhattacharrya distance (current image and reference image from category 6), this is computed for the V or value channel.
- f20: Histogram comparison using the Bhattacharrya distance (current image and reference image from category 7), this is computed for the V or value channel.

The histograms of features f5 to f20 are computed using the Bhattacharrya distance [3, 8]. If we consider two normalized histograms $P$ and $Q$ extracted from an image of size $N \times M$, with $b_i$ (respectively, $q_i$) the $i^{th}$ bins of $P$ (respectively, $Q$), $i = 1, ..., B$, the Bhattacharrya distance $d^{bt}$ between $P$ and $Q$ is given by Eq. 4:

$$d^{dt} = -log \sum_{i=1} \sqrt{P(b_i)Q(q_i)} \tag{4}$$

At last, features f21 to f24 are computed using the GLCM and the two previously described homogeneity and entropy, as shown below:

- f21: GLCM homogeneity ($\Delta x = 1, \Delta y = 0$), 0 degrees.
- f22: GLCM homogeneity ($\Delta x = 0, \Delta y = 1$), 90 degrees.
- f23: GLCM entropy ($\Delta x = 1, \Delta y = 0$), 0 degrees.
- f24: GLCM entropy ($\Delta x = 0, \Delta y = 1$), 90 degrees.

## 4 Experimental results

This section covers two types of experiments using (1) unsupervised clustering algorithms and (2) supervised classification algorithms. Usually, supervised learning is more robust than unsupervised learning. This is due to the extra training phase and access to the label or class (in our case, the categories of emerald stones).

However, we still report experiments using unsupervised approaches as our dataset does not contain all the possible stone categories, and therefore unsupervised learning algorithms such as $k$-Means are straightforwardly adapted to consider more classes. In fact, no extra step is required in this case. If novel classes of stones are introduced to the problem, the $k$-Means algorithm is able to behave normally and separate instances as expected. However, when it comes to supervised learning algorithms, the algorithms require retraining as soon as new classes (e.g., new/different stone categories) are introduced into the problem.

### 4.1 Unsupervised clustering algorithms

The first categorization results are obtained using the k-Means algorithm [26], which is an unsupervised clustering algorithm. Table 3 shows the results obtained with k-Means. Incorrect groupings are highlighted in Italic. A total of 17 stones out of 192 were incorrectly grouped (91.32% correctly grouped instances).

Most incorrect clusterizations narrow down to two main scenarios. The first scenario is related to the darker emerald stones, where 9 instances from category 5 were grouped in category 2. Figure 6a shows two stones that were grouped in the same cluster. These two stones are very similar when analyzed by the naked eye. The stone in Fig. 6b is brighter than that in Fig. 6a. However, there is a fine line that separates both of these images. This type of mistake (or even worse ones) also occurs in the usual manual stone grading processes.

Incorrect clusterizations also happened with brighter stones (categories 0, 3 and 6). Figure 7 shows an error that occurred between categories 0 and 3. In this case, the stone in Fig. 7a is brighter than the one in Fig. 7b. It is important to highlight that the stone classified as 0 is one of the less bright emeralds in category 0, being a borderline stone in the group.

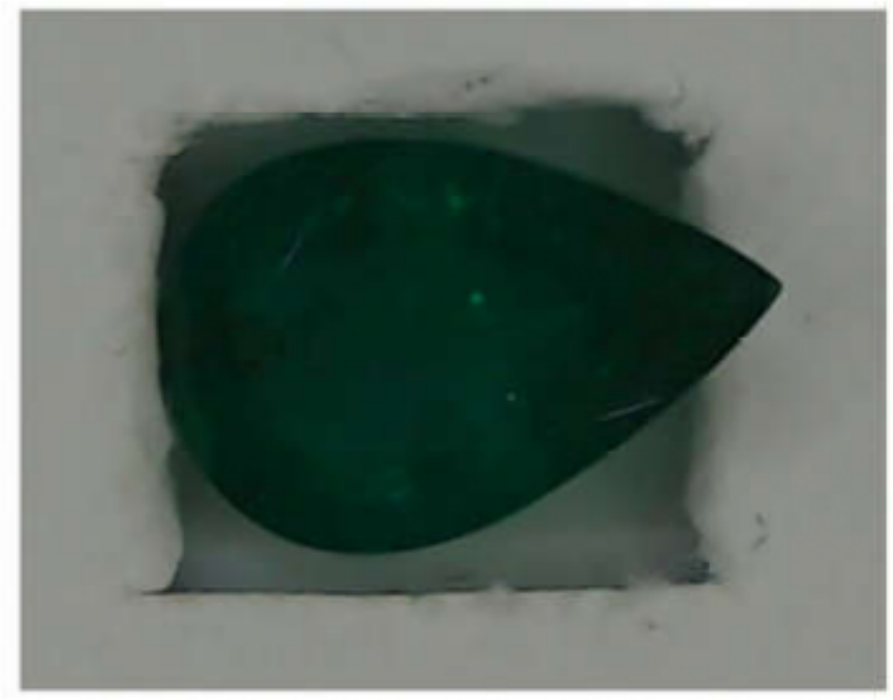

(a) Category 2 clusterized as 5 by *K-Means*.

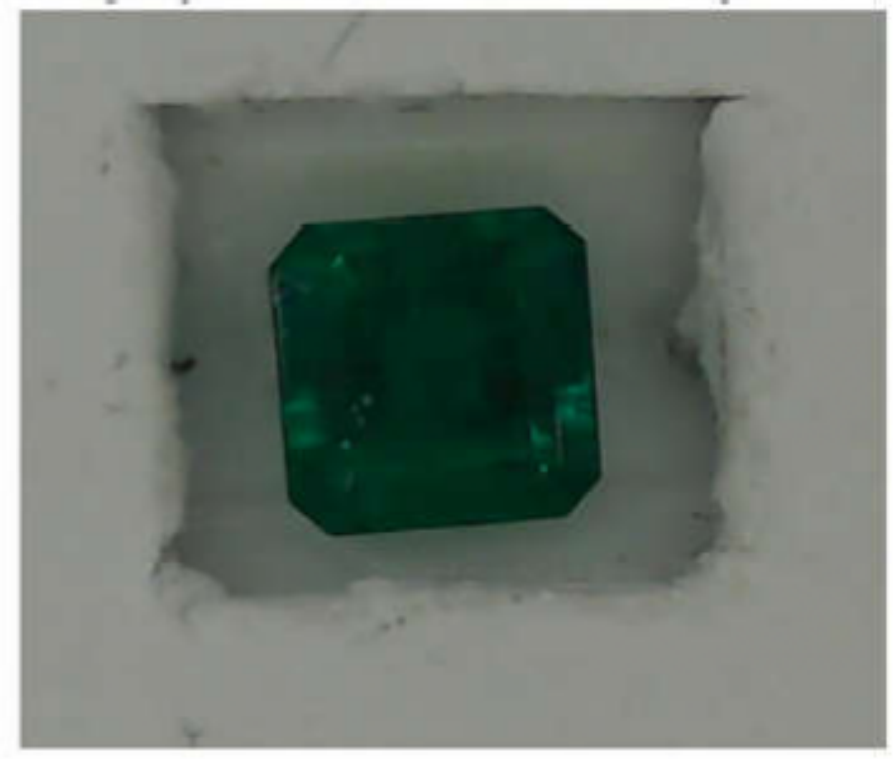

(b) Category 5 correctly clusterized by *K-Means*.

**Fig. 6** Incorrect clusterizations of instances from categories 2 and 5

We also use the *Affinity Propagation* (AF) algorithm and analyze how it compares to k-Means. Results obtained with AF can be seen in Table 4. This clustering algorithm achieved 30 incorrect clusterizations in a total of 192 (84%), being worse than k-Means. Most incorrect clusterizations occurred among categories 2 and 5, where 13 instances from category 5 were clusterized as category 2. It is important to highlight that as AF does not use a $k$ parameter that defines the number of clusters, the algorithm created a total of 8 clusters, as shown in Table 4.

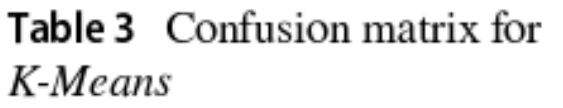

**Table 3** Confusion matrix for *K-Means*

| | Chosen cluster | | | | | | | |
|---|---|---|---|---|---|---|---|---|
| | 0 | 1 | 2 | 3 | 4 | 5 | 6 | 7 |
| Category 0 | 18 | 0 | 0 | *2* | 0 | 0 | *4* | 0 |
| Category 1 | 0 | 23 | 0 | 0 | 0 | 0 | 0 | *1* |
| Category 2 | 0 | 0 | 24 | 0 | 0 | 0 | 0 | 0 |
| Category 3 | 0 | 0 | 0 | 24 | 0 | 0 | 0 | 0 |
| Category 4 | 0 | 0 | 0 | 0 | 24 | 0 | 0 | 0 |
| Category 5 | 0 | 0 | *9* | 0 | 0 | 15 | 0 | 0 |
| Category 6 | 0 | 0 | 0 | *1* | 0 | 0 | 23 | 0 |
| Category 7 | 0 | 0 | 0 | 0 | 0 | 0 | 0 | 24 |

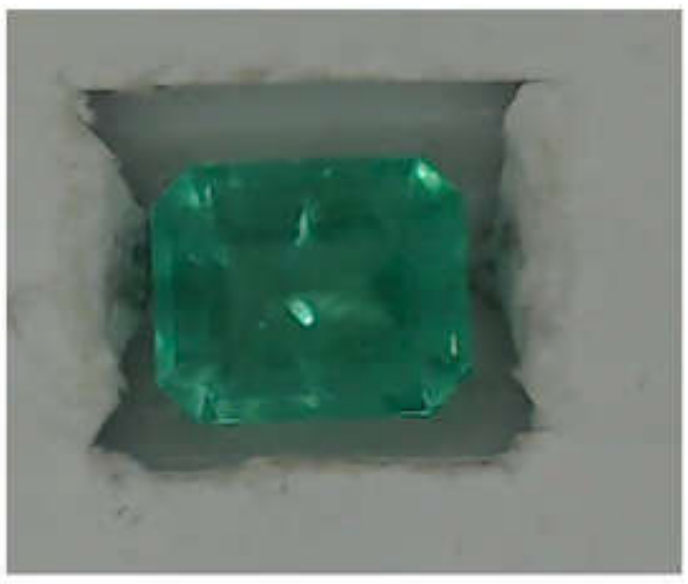

(a) Stone from category 0 being clusterized as 3 by *k-Means*.

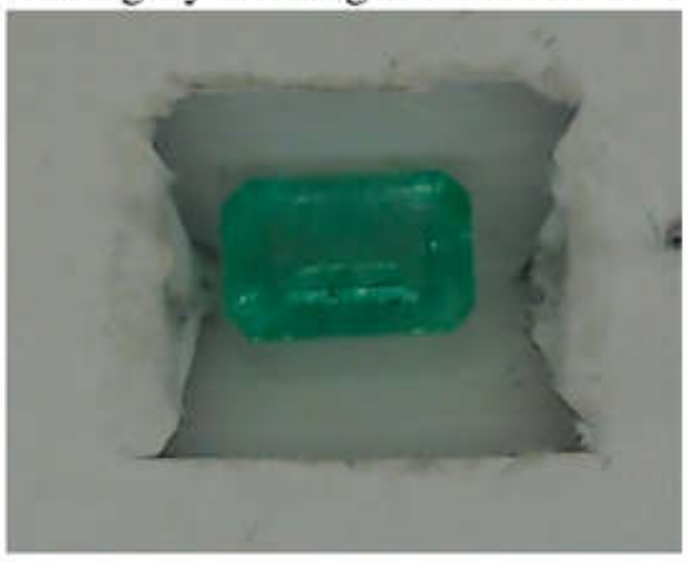

(b) Category 3 being correctly classified.

**Fig. 7** Incorrect clusterizations of instances belonging to categories 0 and 3

As a remark, we also experimented with a hierarchical clustering algorithm called agglomerative hierarchical clustering. However, this approach correctly clusterized just 83.85% of the instances. Hierarchical clustering approaches are a different class of unsupervised clustering algorithms that usually relate more to density, which is not the case for this work and justifies the lower score.

## 4.2 Supervised classification algorithms

In this section, we describe the usage of classification algorithms. As previously mentioned, supervised algorithms such as classifiers are usually better in terms of accuracy (i.e., correctly classified instances) than unsupervised ones. However, they require an additional training phase. We consider three different algorithms: Multilayer Perceptron (MLP), *Random Forests* and *Extremely Randomized Tree* for this set of experiments. All the supervised experiments were evaluated using the fourfold cross-validation method.

The scikit-learn library was used to run the classical machine learning algorithms (MLP, random forests and extremely randomized tree). The parameters of the MLP classifier are given as follows: (hidden-layer-sizes=0.25, solver=sgd, learning-rate=0.01). For random forests and extremely randomized tree, we used the following set of parameters: (cross-validation=4, scoring=f1-weighted, max-depth=5).

Accuracy is defined in Eq. 5 and represents the rate of the correctly classified instances, where TP is the true positive, TN is the true negative, FP is the false positive and FN is the false negative rates.

$$Accuracy = \frac{Correctly\ classified\ instances\ (TP + TN)}{Total\ population\ (TP + TN + FP + FN)} \quad (5)$$

We also provide the f1-score, which is usually used in the context of unbalanced datasets. F1-score is also known as the dice index. The f1-score is defined in Eq. 6.

$$F1score = \frac{TP}{TP + \frac{FP+FN}{2}} \quad (6)$$

### 4.2.1 MLP

Figure 8 shows the results obtained with MLP. The classifier achieved up to 100% of accuracy in the training phase and 98% in the validation, which is a positive aspect. It is important to highlight that the training and validation curves are close to each other over all the iterations. An overfitting to the training data is usually configured when these two curves start to distance substantially from each other.

Figures 9 and 10 show the training and validation accuracy of *Random Forests* and *Extremely Randomized Tree* algorithms. In a similar fashion, training and validation

**Table 4** Confusion matrix for the *Affinity Propagation* algorithm

| | Chosen clusters | | | | | | | | |
|---|---|---|---|---|---|---|---|---|---|
| | 0 | 1 | 2 | 3 | 4 | 5 | 6 | 7 | 8 |
| Category 0 | 19 | 0 | 0 | *1* | 0 | 0 | **4** | 0 | 0 |
| Category 1 | 0 | 23 | 0 | 0 | 0 | 0 | 0 | **1** | 0 |
| Category 2 | 0 | 0 | 23 | 0 | **1** | 0 | 0 | 0 | 0 |
| Category 3 | 0 | 0 | 0 | 24 | 0 | 0 | 0 | 0 | 0 |
| Category 4 | 0 | 0 | 0 | 0 | 24 | 0 | 0 | 0 | 0 |
| Category 5 | 0 | 0 | **13** | 0 | **1** | 10 | 0 | 0 | 0 |
| Category 6 | 0 | 0 | 0 | 0 | 0 | 0 | 15 | 0 | **9** |
| Category 7 | 0 | 0 | 0 | 0 | 0 | 0 | 0 | 24 | 0 |

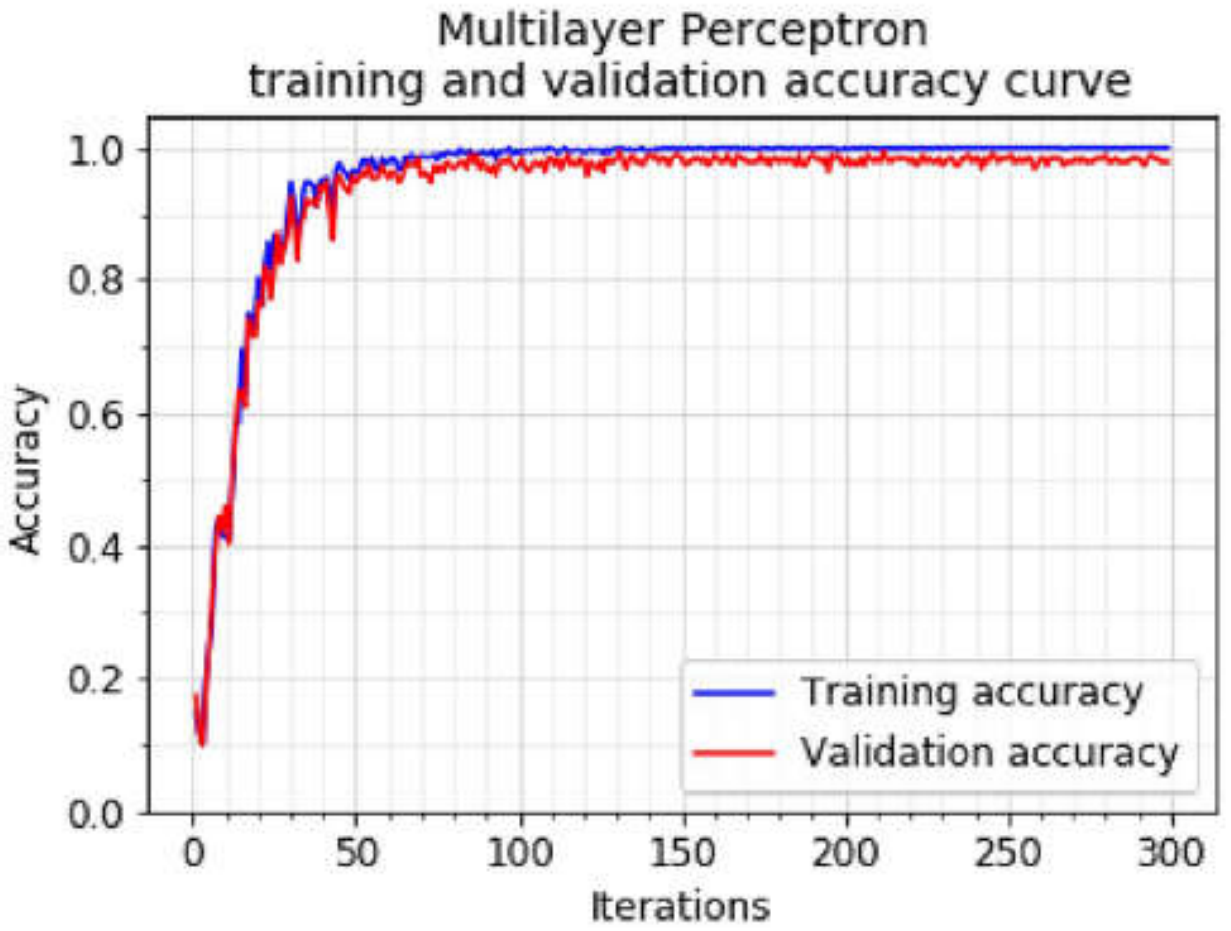

**Fig. 8** Results obtained with the MLP algorithm

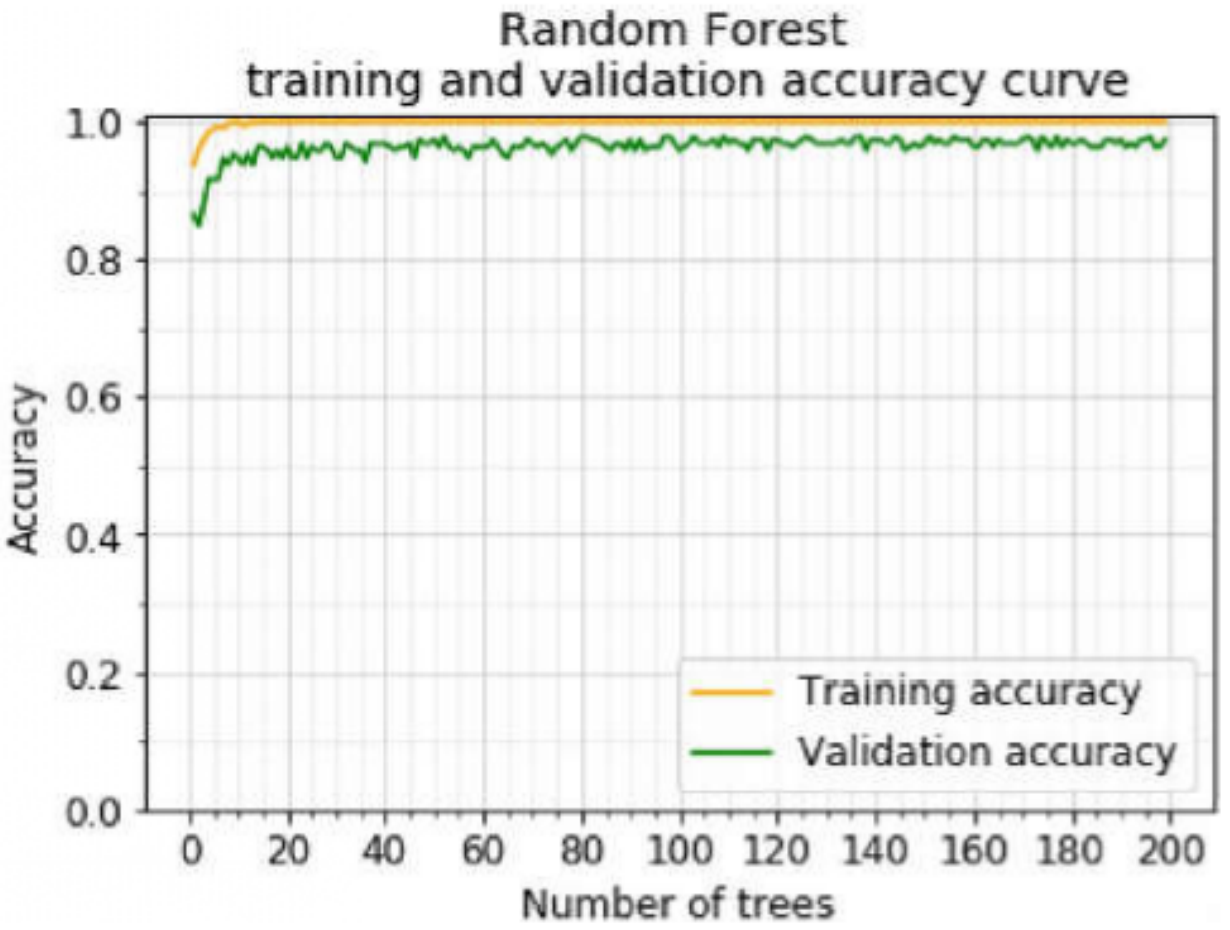

**Fig. 9** Results obtained with *Random Forests*

curves are close to each other, which does not indicate an overfitting bias.

Table 5 compares the results obtained with all algorithms including both supervised and unsupervised paradigms. Supervised approaches achieved better results, as expected. However, as previously mentioned, unsupervised algorithms are amenable to the seamlessly insertion of new classes into the problem.

A very important remark is that classification and clusterization errors, overall, are related to the same categories or classes over all the approaches and paradigms.

Supervised learning obtained exceptional results, up to 98% of accuracy with MLP, rightly followed by random forests and extremely randomized tree. This translates to just four stones being incorrectly classified out of 192. In this sense, this is a very efficient framework for the categorization of emerald stones.

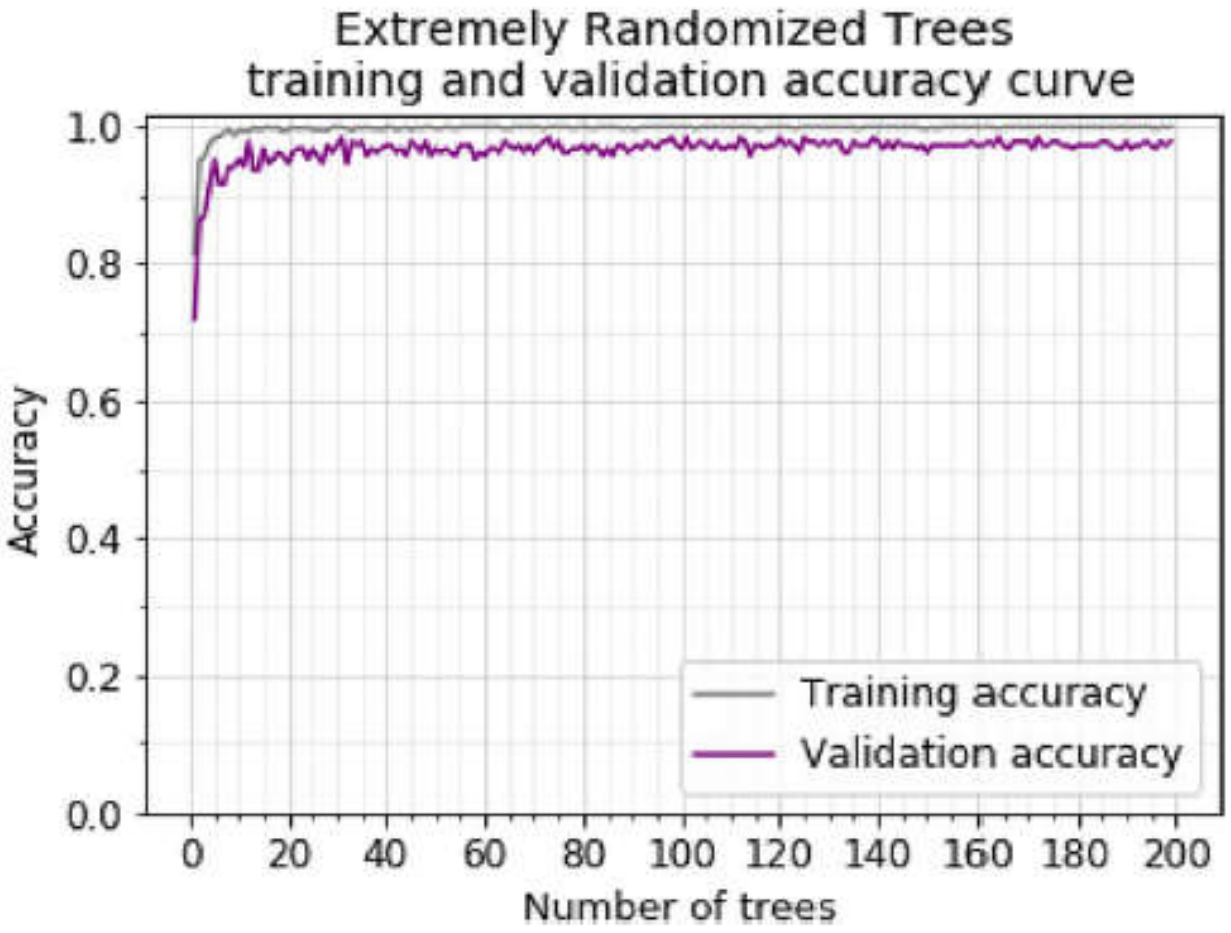

**Fig. 10** Results obtained with the *Extremely Randomized Tree* algorithm

**Table 5** Table containing the accuracy and f1-score

| Unsupervised | Accuracy | F1-score |
|---|---|---|
| *Affinity propagation* | 86.0% | 0.8639 |
| K-Means | 91.0% | 0.8843 |
| *Supervised* | | |
| MLP | 98.0% | 0.9842 |
| *Extremely randomized tree* | 97.0% | 0.9793 |
| *Random forest* | 97.0% | 0.9737 |

In addition, we also analyzed the performance of the Nvidia Digits framework, which is a deep learning framework that works using the GPU. The deep learning approach achieved 89% of accuracy. This rate was obtained using 256x256 images as input (resized automatically by the digits framework), the AlexNet network [17], and small variations on the learning rate, where the chosen learning rate was set to 0.001. A total of 144 images were used for training and 44 images for validation (hold out). The network was trained for two days using an i5-9400F processor and the Nvidia GTX 1660.

Deep learning is currently the state-of-the-art in terms of image recognition. However, our classical approach was able to achieve better results. We strongly believe that this result is associated with the usage of the features based on reference images (expert knowledge), which is something that a convolutional neural network/deep learning cannot infer directly (features f5 to f20 entail this information). These reference images can be found in the following address: [6]. At last, the size of the dataset may also influence the AlexNet result. Larger quantities of emerald images or data augmentation could potentially improve the results.

### 4.3 Feature ranking

We will now shift our attention toward a feature ranking generated by the Weka framework [10, 14]. The rank was created using the Classifier Attribute Evaluator algorithm associated with the random forests and multilayer perceptron algorithms, which is shown in Table 6. The Classifier Attribute Evaluator removes and reinserts attributes into the dataset and analyses how their presence influences the obtained accuracy of RF and MLP. When the inclusion of certain attributes increases the accuracy rate of the classifier, these attributes receive better overall positions in the rank.

**Table 6** Feature ranking using the Weka classifier attribute evaluator

| Rank | Random forests | Multilayer perceptron |
|---|---|---|
| 1 (Best) | f3: arithmetic mean (channel V) | f3: arithmetic mean (channel V) |
| 2 | f9: histogram comparison (reference image from category 4, channel S) | f12: histogram comparison (reference image from category 7, channel S) |
| 3 | f5: histogram comparison (reference image from category 0, channel S) | f9: histogram comparison (reference image from category 4, channel S) |
| 4 | f12: histogram comparison (reference image from category 7, channel S) | f5: histogram comparison (reference image from category 0, channel S) |
| 5 | f10: histogram comparison (reference image from category 5, channel S) | f10: histogram comparison (reference image from category 5, channel S) |
| 6 | f1: arithmetic mean (channel S) | f8: histogram comparison (reference image from category 3, channel S) |
| 7 | f8: histogram comparison (reference image from category 3, channel S) | f1: arithmetic mean (channel S) |
| 8 | f11: histogram comparison (reference image from category 6, channel S) | f11: histogram comparison (reference image from category 6, channel S) |
| 9 | f14: histogram comparison (reference image from category 1, channel V) | f6: histogram comparison (reference image from category 1, channel S) |
| 10 | f15: histogram comparison (reference image from category 2, channel V) | f20: histogram comparison (reference image from category 7, channel V) |
| 11 | f6: histogram comparison (reference image from category 1, channel S) | f22: GLCM homogeneity ($\Delta x = 0, \Delta y = 1$) |
| 12 | f17: histogram comparison (reference image from category 4, channel V) | f21: GLCM homogeneity ($\Delta x = 1, \Delta y = 0$) |
| 13 | f22: GLCM homogeneity ($\Delta x = 0, \Delta y = 1$) | f7: histogram comparison (reference image from category 2, channel S) |
| 14 | f7: histogram comparison (reference image from category 2, channel S) | f14: histogram comparison (reference image from category 1, channel V) |
| 15 | f19: histogram comparison (reference image from category 6, channel S) | f17: histogram comparison (reference image from category 4, channel V) |
| 16 | f23: GLCM entropy ($\Delta x = 1, \Delta y = 0$) | f24: GLCM entropy ($\Delta x = 0, \Delta y = 1$) |
| 17 | f21: GLCM homogeneity ($\Delta x = 1, \Delta y = 0$) | f23: GLCM entropy ($\Delta x = 1, \Delta y = 0$) |
| 18 | f20: histogram comparison (reference image from category 7, channel V) | f2: standard deviation (S channel) |
| 19 | f24: GLCM entropy ($\Delta x = 0, \Delta y = 1$) | f15: histogram comparison (reference image from category 2, channel V) |
| 20 | f16: histogram comparison (reference image from category 3, channel V) | f16: histogram comparison (reference image from category 3, channel V) |
| 21 | f2: standard deviation (channel S) | f4: standard deviation (channel S) |
| 22 | f18: histogram comparison (reference image from category 5, channel V) | f15: histogram comparison (reference image from category 2, channel V) |
| 23 | f4: standard deviation (channel V) | f16: histogram comparison (reference image from category 3, channel V) |
| 24 | f13: histogram comparison (reference image from category 0, channel V) | f4: standard deviation (channel V) |

Best attributes are placed at the top of the ranking. It was possible to observe in the result logs that all the features obtained positive scores, which indicates that all the chosen features are important for the classification. However, individually, Table 6 highlights that the features based on the histogram using the reference images are usually in the top of the ranking.

Besides the Classifier Attribute Evaluator, we also evaluated other three types of attribute selection algorithms. One based on the information gain, gain ratio (information gain normalized by the entropy of the attribute) and correlation, all with respect to the class (respectively called InfoGainAttributeEval, GainRatioAttributeEval and CorrelationAttributeEval in the Weka framework).

Information gain and gain ratio generate almost the same rank, which is similar to the one shown in Table 6. Features were ranked as follows according to the information gain approach (top 10): f3, f9, f12, f1, f10, f5, f8, f11, f6, f20. The correlation, on the other hand, generates a substantially different rank, as follows, in respect to the feature number (top 10): f14, f16, f15, f17, f1, f12, f3, f5, f10, f8.

The Classifier Attribute Evaluator measures the importance of the feature on the classification accuracy. This method is especially important because it already works with the classifier that is being used to classify the stones. This fact guarantees an improvement on the classification performance. However, care should be taken to avoid overfitting. Algorithms that are less affected by overfitting such as RF are better choices.

The feature ranking is useful to measure the importance of the feature individually, as well as to reduce the dimensionality of the problem. However, for this specific problem, when we reduced the size of the feature set, we obtained worse results. RF reached 85.41% of accuracy using the top 5 features and 92.70% of accuracy using the top 10. This also reinforces the fact that all the chosen features are valuable as they achieved positive scores in the attribute evaluator algorithm. This indicates that these features do not disturb the performance of the classifier, which can often happen in other scenarios due to the curse of dimensionality, for instance.

## 5 Conclusion and future work

Pricing, grading and categorization of emerald stones are currently a manual procedure performed by specialists on gemology. A popular approach is to compare the stone to a reference stone to infer its grading, which is a very subjective process. Two different observers may have slight different pattern strategies in terms of the most similar reference stone. Several other external factors also influence on this decision, as previously described.

In order to overcome this situation and to provide a methodology that enables seamless automation, we propose a framework that uses image processing techniques coupled with machine learning. This work is the very first in the literature to propose an automated approach for the classification and categorization of emerald stones.

Accuracies were as high as 98% using the multilayer perceptrons classifier, which is a good accuracy rate. The proposed framework was more efficient than a deep learning approach (Nvidia Digits). Unsupervised algorithms such as k-Means achieved similar but slighly worse results when compared to supervised approaches, up to 91% of correctly clusterized stones.

As a final remark, we also created the dataset containing the emerald stones used in this work, which is made public and is available for further comparison, reproduction and study at the following repository: [6].